\documentclass[conference]{IEEEtran}

\usepackage{hyperref}
\usepackage{url}

\usepackage{balance}
\usepackage{graphicx}
\usepackage{caption}
\usepackage{subcaption}
\usepackage{wrapfig}
\usepackage[dvipsnames]{xcolor}
\usepackage{amsmath,amssymb,amsfonts}
\usepackage{algorithm}
\usepackage{makecell}
\usepackage{algpseudocode}
\usepackage{tikz}
\usepackage{pgfplots}
\usepackage{calc}  
\usepackage{enumitem} 
\usepackage{amsmath}
\usepackage{xcolor}
\usepackage{booktabs}
\usepackage{multirow}
\usepackage{amsthm} 
\usepackage[capitalize,noabbrev]{cleveref}
\usepackage{soul}

\definecolor{codegreen}{rgb}{0,0.6,0}
\definecolor{codegray}{rgb}{0.5,0.5,0.5}
\definecolor{codepurple}{rgb}{0.58,0,0.82}
\definecolor{backcolour}{rgb}{0.95,0.95,0.92}
\definecolor{jcred}{HTML}{e31a1c}
\definecolor{jcgreen}{HTML}{33a02c}
\definecolor{jcblue}{HTML}{1f78b4}
\definecolor{jcorange}{HTML}{ff7f00}
\definecolor{jcpurple}{HTML}{6a3d9a}
\definecolor{jclightred}{HTML}{fb8072}
\definecolor{jclightgreen}{HTML}{b3de69}
\definecolor{jclightblue}{HTML}{80b1d3}
\definecolor{jclightorange}{HTML}{fdb462}
\definecolor{jclightpurple}{HTML}{bebada}
\definecolor{jcredl}{HTML}{fb8072}
\definecolor{jcgreenl}{HTML}{b3de69}
\definecolor{jcbluel}{HTML}{80b1d3}
\definecolor{jcorangel}{HTML}{fdb462}
\definecolor{jcpurplel}{HTML}{bebada}
\definecolor{jcbluem}{HTML}{488bb8}
\definecolor{jcyellow}{HTML}{ffff99}
\definecolor{jcbrown}{HTML}{b15928}

\def\BibTeX{{\rm B\kern-.05em{\sc i\kern-.025em b}\kern-.08em
    T\kern-.1667em\lower.7ex\hbox{E}\kern-.125emX}}

\newcommand*\best[1]{\textcolor{jcgreen}{\bf #1}}

\title{LLM4DV: Using Large Language Models for
Hardware Test Stimuli Generation}

\author{%
  \IEEEauthorblockN{
  Zixi Zhang\textsuperscript{{1}, *}, 
  Balint Szekely\textsuperscript{{2}, *}, 
  Pedro Gimenes\textsuperscript{{2}}, 
  Greg Chadwick\textsuperscript{{3}}, \\ 
  Hugo McNally\textsuperscript{{3}}, 
  Jianyi Cheng\textsuperscript{{4}}, 
  Robert Mullins\textsuperscript{{1}} and
  Yiren Zhao\textsuperscript{{2}}
  }
  \IEEEauthorblockA{\textsuperscript{{1}}University of Cambridge, UK; \textsuperscript{{2}}Imperial College London, UK; \textsuperscript{{3}}lowRISC, UK; \textsuperscript{{4}}University of Edinburgh, UK\\
  \textsuperscript{*} Equal Contribution\\
  Email: \text{zz458@cam.ac.uk, robert.mullins@cl.cam.ac.uk},
  \text{\{balint.szekely20, pedro.gimenes19, a.zhao\}@imperial.ac.uk}, \\
  \text{\{gac, hugom\}@lowrisc.org},
  \text{jianyi.cheng@ed.ac.uk}
  }
}

\begin{document}

\maketitle

\begin{abstract}
Hardware design verification (DV) is a process that checks the functional equivalence of a hardware design against its specifications, improving hardware reliability and robustness. A key task in the DV process is the test stimuli generation, which creates a set of conditions or inputs for testing. These test conditions are often complex and specific to the given hardware design, requiring substantial human engineering effort to optimize. 
We seek a solution of automated and efficient testing for arbitrary hardware designs that takes advantage of large language models (LLMs). LLMs have already shown promising results for improving hardware design automation, but remain under-explored for hardware DV. In this paper, we propose an open-source benchmarking framework named LLM4DV that efficiently orchestrates LLMs for automated hardware test stimuli generation. Our analysis evaluates six different LLMs involving six prompting improvements over eight hardware designs and provides insight for future work on LLMs development for efficient automated DV.
\end{abstract}

%
\IEEEpeerreviewmaketitle

\section{Introduction}
\vspace{-0.5em}

Large Language Models (LLMs)~\cite{yang2020bert, introducing_chatgpt, llama2:2023} have gained significant attention in recent years due to their language generation and comprehension capabilities on tasks such as language translation~\cite{feng2020language}, question answering~\cite{yang2020bert}, and sentiment analysis~\cite{liu2021makes}. 
Recently, there has been interest in exploiting LLMs to improve hardware design generation~\cite{blocklove2023chip, fu2023gpt4aigchip, lu2024rtllm}. Arguably, hardware design verification (DV), which checks the correctness of hardware designs, ranks among the most crucial and time-consuming tasks in hardware development. Hardware DV is often {\bf \em time-consuming}, usually taking up to 60\%-70\% of the development time~\cite{dv_time}, and requires significant {\bf \em human guidance and expertise} due to the complexity of both hardware design and its corresponding testing requirements~\cite{shin2022data}.

On the other hand, existing work on LLMs has been studied for software testing. For example, Codex \cite{eval:2021} can produce functionally correct bodies of code from natural language docstring descriptions. LLaMA \cite{llama2:2023}, an LLM using instruction tuning and Reinforcement Learning with Human Feedback (RLHF) \cite{human_pref:2017, sum_human_feedback:2020} for fine-tuning, emerges impressive generalization and external tool usage ability. However, these approaches are not directly applicable due to the following two challenges. First, unlike software programming languages, there is a scarcity of high-quality, open-source hardware designs and testing code available online for training LLMs. This limitation is critical because Hardware Description Languages (HDLs) possess {\bf \em distinct semantics} that differ fundamentally from software programming languages. These unique characteristics make HDLs considerably more challenging for LLMs to interpret and learn from, as the models cannot simply transfer their knowledge from conventional programming contexts without substantial modifications. Second, the testing space for a hardware design design is typically large, leading to a {\bf \em scalability} problem. Existing approaches on hardware DV require human guidance to reduce search space, such as adding heuristics to guide tests of a particular hardware design. This raises an important question: {\bf \em can LLMs effectively minimize the amount of human effort involved in hardware DV?} 


\begin{figure}
    \centering
    \includegraphics[width=\linewidth]{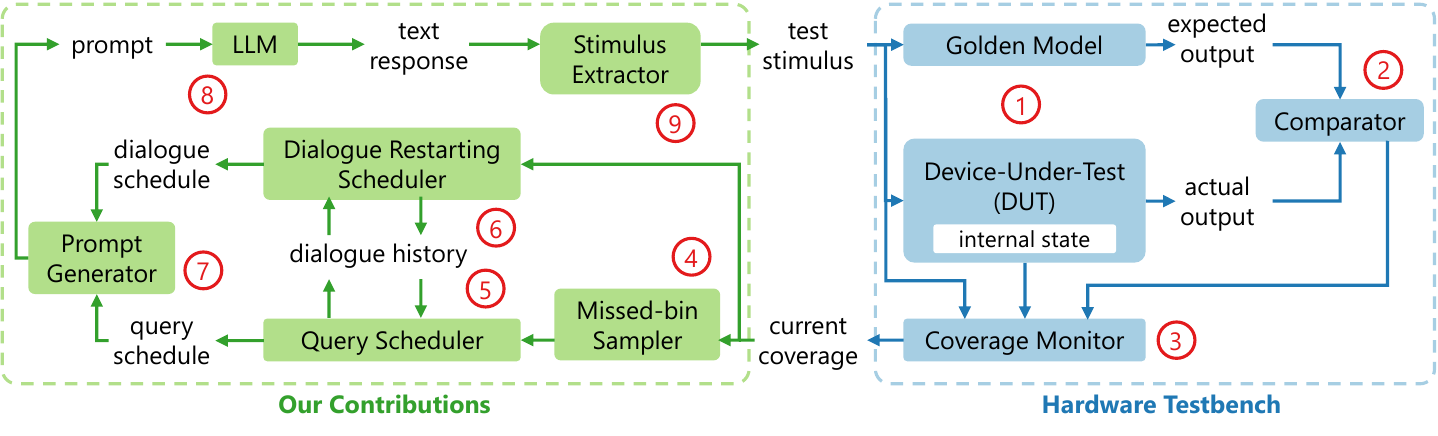}
    \caption{An overview of LLM4DV framework. The {\color{jcblue} \em right part} shows a traditional DV process. DV engineers need to manually interact with the DV process by tailoring various stimulus and observing the coverage. Such a manual process is often iterative. The {\color{jcgreen} \em left part} highlights our contributions, which adds the stimulus generation agent for automated guidance. }
    \label{LLM4DV_extended}
\end{figure}

\begin{table*}
\small
  \caption{Comparison to related work applying LLMs in the field of digital hardware.}
  \label{benchmark_comp}
  \centering
  \resizebox{0.9\textwidth}{!}{
  \begin{tabular}{llcr}
    \toprule
    \multicolumn{1}{c}{\textbf{Name}}     & \multicolumn{1}{c}{\textbf{Task}}  & \multicolumn{1}{c}{\textbf{Number of models}}    & \multicolumn{1}{c}{\textbf{Testing space}} \\
    \midrule
    RTLFixer\cite{rtlfixer:2023} & Verilog syntax correction & 1 & 212 syntax errors   \\
    NSPG\cite{NSPG:2023} & Repairing security-relevant bugs in Verilog & 4 & 10 designs (10 bugs)   \\
    ChipNeMo\cite{chipnemo:2023} & Bug analysis and summarisation & 3 & 30 bugs   \\
    Kande et al.\cite{llm_assisted_assertions:2023} & Generating security assertions & 4 & 10 designs (10 assertions)   \\
    Thakur et al.\cite{thakur:2023} & Generating Verilog code & 6 & 17 problems   \\
    RTLLM \cite{RTLLM:2023} & Generating Verilog code & 4 & 30 designs   \\
    \midrule
    LLM4DV & Stimulus generation for functional verification & 5 & 8 designs (3883 coverage bins)   \\
    \bottomrule
  \end{tabular}
  }
\end{table*}

In this work, we specifically focus on hardware test stimuli generation, which generates test inputs for hardware DV. 
A good stimuli discovers new hardware states during testing, increasing the test coverage; while a bad stimuli only tests existing states, leaving the coverage the same. 
Finding good stimuli becomes particularly arduous when encountering hard-to-hit points within the coverage plan. 
In order to find a path to LLM solutions, we present a novel benchmarking framework named LLM4DV (Large Language Model for Design Verification), that utilizes LLMs for \textit{\textbf{test stimuli generation}}, and make the following contributions:
\begin{itemize}[leftmargin=10pt]
    \item We introduce and construct LLM4DV, a framework that employs prompted LLMs to generate test stimuli for hardware DV. 
    We show automated DV requires a complex prompting strategy and also propose six prompt enhancements to establish strong baselines for the LLM4DV framework. We believe this provides an attractive testbed for experimenting the agentic behavior of LLMs.
    
    \item We design and construct three DUT modules: a Primitive Data Prefetcher Core, an Ibex CPU Instruction Decoder, and an Ibex CPU. We also select five open-source designs, obtaining a set of DUTs with different testing difficulties. We evaluate LLM4DV using these eight DUT modules and introduce a set of evaluation metrics. LLMs, with optimized prompt enhancements, achieve coverage rates (a primary metric for measuring verification effectiveness) ranging from 89.74\% to 100\% in a realistic setting. We open-source LLM4DV, the workflow and our optimized prompts, alongside these modules to allow the community to experiment with their ideas.
\end{itemize}


\section{Background and Related Work}
\label{sec:overview}



A traditional hardware DV process is illustrated in the right of Figure~\ref{LLM4DV_extended}. For each hardware design, also known as device-under-test (DUT), the hardware designer provides a functionally equivalent golden model in software to the DUT~\cite{ACM:2022}. The DV process takes a set of inputs, or test stimuli, and sends them to both the DUT and its golden model ({\textcolor{jcred}{\textcircled{\raisebox{-0.9pt}{1}}}}), leading to two sets of results. The results are then compared between the DUT and its golden model ({\textcolor{jcred}{\textcircled{\raisebox{-0.9pt}{2}}}}). 
A DV process typically tests the DUT iteratively on a large set of stimuli defined by the hardware designer in advance, which is known as the \textbf{\textit{coverage plan}}, and is normally in the form of \textbf{\textit{coverage bins}}. 
A coverage bin is a specific condition or scenario that the verification environment tracks to determine whether a particular aspect of the design has been exercised or tested. 
The coverage monitor ({\textcolor{jcred}{\textcircled{\raisebox{-0.9pt}{3}}}}) inspects the DUT’s inputs, outputs, and internal states; determines whether there are hits of coverage bins; updates the current coverage and returns it to the stimulus generation agent for the next stimulus. The procedure in the right of Figure~\ref{LLM4DV_extended} typically follows an iterative approach, often executed tens of thousands of times, in which a human DV engineer applies various stimuli to achieve comprehensive coverage specified in the coverage plan.


Effective test stimuli generation has been a major challenge especially if one wants to meet 100\% coverage~\cite{ACM:2022}. For a simple design, verification can be done with individual directed tests, in which test stimuli (inputs for the DUT) are manually generated. For more complex designs, a large number of stimuli is required for exercising as much of the design’s functionality as possible. Traditionally, \textbf{\textit{constrained-random testing (CRT)}} has been used to generate vast random but valid test stimuli and to attempt to ``hit'' the bins. However, CRT is inefficient to hit as many bins as human effort for hardware states with complicated conditions. 

In hardware design verification, assertion-based verification (ABV) is also widely adopted. ABV inserts assertions into the DUT HDL source to detect violations of predefined design properties. However, ABV requires test patterns (i.e. input test stimuli) to activate given assertions and therefore reveal vulnerabilities. For simulation-based ABV approaches, traditional test generation that uses random or constrained-random tests cannot guarantee to activate assertions with complex conditions in a reasonable time, even with optimizations \cite{adaptive_testbenches:2008, gen_tlm:2008, airwolf:2009, activation_assertion:2020}, and can face the complexity explosion problems \cite{ACM:2022}. 
We present an alternative solution ito this issue by utilizing LLM’s pre-trained knowledge to reason about the given coverage plan and guide the test stimuli generation. Other advanced testing techniques, such as coverage-directed generation and mutating tests~\cite{fine2003coverage, guzey2007coverage, laeufer2018rfuzz}, have been studied to improve the performance of CRT. 

Recently, the application of LLMs for hardware design and verification purposes has started to gain traction \cite{LLM4EDA:2023}. Table \ref{benchmark_comp} provides a summary of recent benchmarks that focus on applying LLMs within this domain. In particular, \textit{there are currently no benchmarks that evaluate the stimuli generation capabilities of LLMs.}
Among the recent contributions, RTLFixer \cite{rtlfixer:2023} enables the automated correction of Verilog syntax errors. In contrast, NSPG \cite{NSPG:2023} is designed to extract security properties by analyzing hardware documentation. ChipNeMo \cite{chipnemo:2023} has been assessed on tasks related to bug summarization and analysis. Additionally, Kande et al. \cite{llm_assisted_assertions:2023} proposed a methodology for automatically generating SystemVerilog assertions (SVAs) using LLMs to enhance hardware security. Meanwhile, Thakur et al. \cite{thakur:2023} and RTLLM \cite{RTLLM:2023} have explored the generation of Register Transfer Level (RTL) code using LLMs. While it is challenging to directly compare the scale of these benchmarks with that of LLM4DV due to the different abilities assessed, it should be noted that LLM4DV's scope of 3883 coverage bins across 8 devices, tested with six different off-the-shelf models, represents a significant contribution to this emerging field.

\begin{table}
\scriptsize
  \caption{A list of input options and output evaluation metrics for the proposed LLM4DV framework.}
  \label{metrics}
  \centering
  \resizebox{\linewidth}{!}{
\begin{tabular}{clp{5cm}}
\toprule
\textbf{} & {\bf Names} & {\bf Descriptions} \\
\midrule
\multirow{4}{*}{\begin{tabular}[c]{@{}c@{}}Input\\Options\end{tabular}} & DUT & The target DUT to be tested. \\
 & Model & The LLM used for stimulus generation. \\
 & Prompting Configurations & The prompting strategy for stimulus generation. \\
 & Coverage Plan & The coverage plan specified for the target DUT. \\
 \midrule
\multirow{6}{*}{\begin{tabular}[c]{@{}c@{}}Evaluation\\ Metrics\end{tabular}} & Max Coverage & Maximum recorded number of coverage bins (defined by the coverage plan) covered. A higher number indicates better performance. \\
 & Effective Message Count & The minimum number of messages an LLM produces across several trials in an experiment achieving maximum coverage; a lower count indicates better performance.
 \\
 & Average Message Count & Average number of query messages per experiment $\pm$ standard deviation of messages. As the usage of LLMs is costly, a faster convergence to maximum coverage is preferred. \\
 \bottomrule
\end{tabular}
}
\end{table}


\section{LLM4DV Benchmarks}
\label{sec:method}


\subsection{LLM4DV Framework}
\label{sec:method:llm4dv}
In this work, the proposed LLM4DV framework automates the DV process by exploiting LLMs for test stimuli generation, shown in the left of Figure~\ref{LLM4DV_extended}. This reduces human involvement in the hardware DV loop and effectively guides tests to increase coverage rates. 
In each generation cycle, the prompt generator produces a prompt based on a template ({\textcolor{jcred}{\textcircled{\raisebox{-0.9pt}{7}}}}) and the current coverage feedback from the coverage monitor. 
LLM4DV allows customization of prompts inside a dialogue, this means each \textbf{\textit{query message}} can receive different prompts, as managed by the query scheduler ({\textcolor{jcred}{\textcircled{\raisebox{-0.9pt}{5}}}}) shown in Figure \ref{LLM4DV_extended}. This is explained in Section~\ref{sec:prompt}.

The LLM takes in the prompt and generates a natural language response, from which the test stimulus values are extracted and sent to the DV flow in the right of Figure~\ref{LLM4DV_extended}. The DV framework then produces current coverage which is considered as input for the LLM-based stimulus generation agent ({\textcolor{jcred}{\textcircled{\raisebox{-0.9pt}{8}}}}) shown in Figure~\ref{LLM4DV_extended}. The processes of test stimuli generation and hardware testbench simulation are executed in parallel asynchronously. A buffer is placed between the stimulus generation agent interfaces to balance the rate of the test stimuli generation and consumption. In every timestep when the stimulus generation agent is requested for a test stimulus, it takes out the oldest value in its buffer; if the buffer is empty, the agent takes in a new request and generates new stimuli.

In LLM4DV, each DV process is viewed as a ``\textbf{\textit{trial}}'', where there would be multiple dialogues made in a single trial, as illustrated in Figure \ref{LLM4DV_extended}, which are controlled by the dialogue scheduler ({\textcolor{jcred}{\textcircled{\raisebox{-0.9pt}{6}}}}).
A \textit{trial} stops if certain user-set conditions are met (eg. max token counts reached), and the agent is considered ``exhausted''. 

\subsection{Evaluation setup: DUTs and models}

The proposed LLM4DV benchmark contains eight DUT modules. 
Three of the devices were developed by the authors, and the other five are open-source designs. These DUTs are selected because they are commonly seen in most representative computer architectures such as CPUs, GPUs and other hardware accelerators. They are:
\begin{itemize}
    \item \textit{Primitive Data Prefetcher Core}: Detects stride patterns in a series of integers.
    \item \textit{Ibex Instruction Decoder}: Decodes RISC-V instructions.
    \item \textit{Ibex CPU}: A RISC-V CPU core.
    \item \textit{Asynchronous FIFO} \cite{async_fifo}: A dual clock FIFO.
    \item \textit{AMPLE Prefetcher Weight Bank} \cite{ample}: A component of AMPLE, a graph neural network accelerator. It is responsible for fetching data from memory, storing it in a FIFO.
    \item \textit{AMPLE Prefetcher Fetch Tag} \cite{ample}: Another component of AMPLE. Similar to the weight bank, its basic purpose is to fetch data from memory. 
    \item \textit{SDRAM Controller} \cite{sdram_controller}: a simple SDRAM controller.
    \item \textit{MIPS CPU} \cite{mips_cpu}: A MIPS CPU core.
\end{itemize}

We use six different commercially available LLMs: GPT-3.5 Turbo, Llama v2 70B Chat, Claude 3 Sonnet, CodeLlama 70B Instruct, Llama 3 70B Instruct, and Claude 3.5 Sonnet. 
To evaluate the effectiveness of these LLMs, we observe the testing performance based on three evaluation metrics, as listed in the lower part of Table~\ref{metrics}. We have limited each trial to 700 messages. The design choices of all parameters and prompting are based on our ablation runs.


\subsection{LLM4DV prompting enhancements}
\label{sec:prompt}

LLM4DV supports custom user prompts at various interaction stages.
We present the following prompting strategies that are already integrated in LLM4DV. These prompting techniques have been empirically observed to influence final performance to varying extents.

\textbf{Missed-bin sampling} This optimization is {\textcolor{jcred}{\textcircled{\raisebox{-0.9pt}{4}}}} in Figure \ref{LLM4DV_extended}, and is later used in the query scheduler. 
We propose missed-bin sampling, which samples a number of bins from all uncovered bins to be included in the differences part of iterative queries. 

\textbf{Best-iterative-message sampling} The LLM needs previous query messages to learn about what has happened. However, as the dialogue grows, the length of input may exceed the LLM’s input limit. 
We propose to randomly keep recent messages and ``successful'' previous messages that hit the largest number of bins.
These strategies are used in our Query Scheduler in {\textcolor{jcred}{\textcircled{\raisebox{-0.9pt}{5}}}}.

\textbf{Dialogue restarting} LLMs sometimes behave stubbornly, repeating mistakes they made previously. To overcome this behavior, we introduce a dialogue restarting scheduler ({\textcolor{jcred}{\textcircled{\raisebox{-0.9pt}{6}}}}). When the LLM hits less than three new bins within $t$ responses, we clear the dialogue record and restart from the system message and initial query.

\textbf{Best-iterative-message buffer reset} When the dialogue record is reset, the buffer for ``successful'' previous messages in \textit{Best-iterative-message sampling} can also be cleared or kept. These two strategies display a trade-off between ``effectively forgetting past mistakes'' and ``faster task re-learning after restart''. This is incorporated in the restarting scheduler ({\textcolor{jcred}{\textcircled{\raisebox{-0.9pt}{6}}}}).




\begin{figure}
  \centering
  \includegraphics[width=0.9\linewidth]{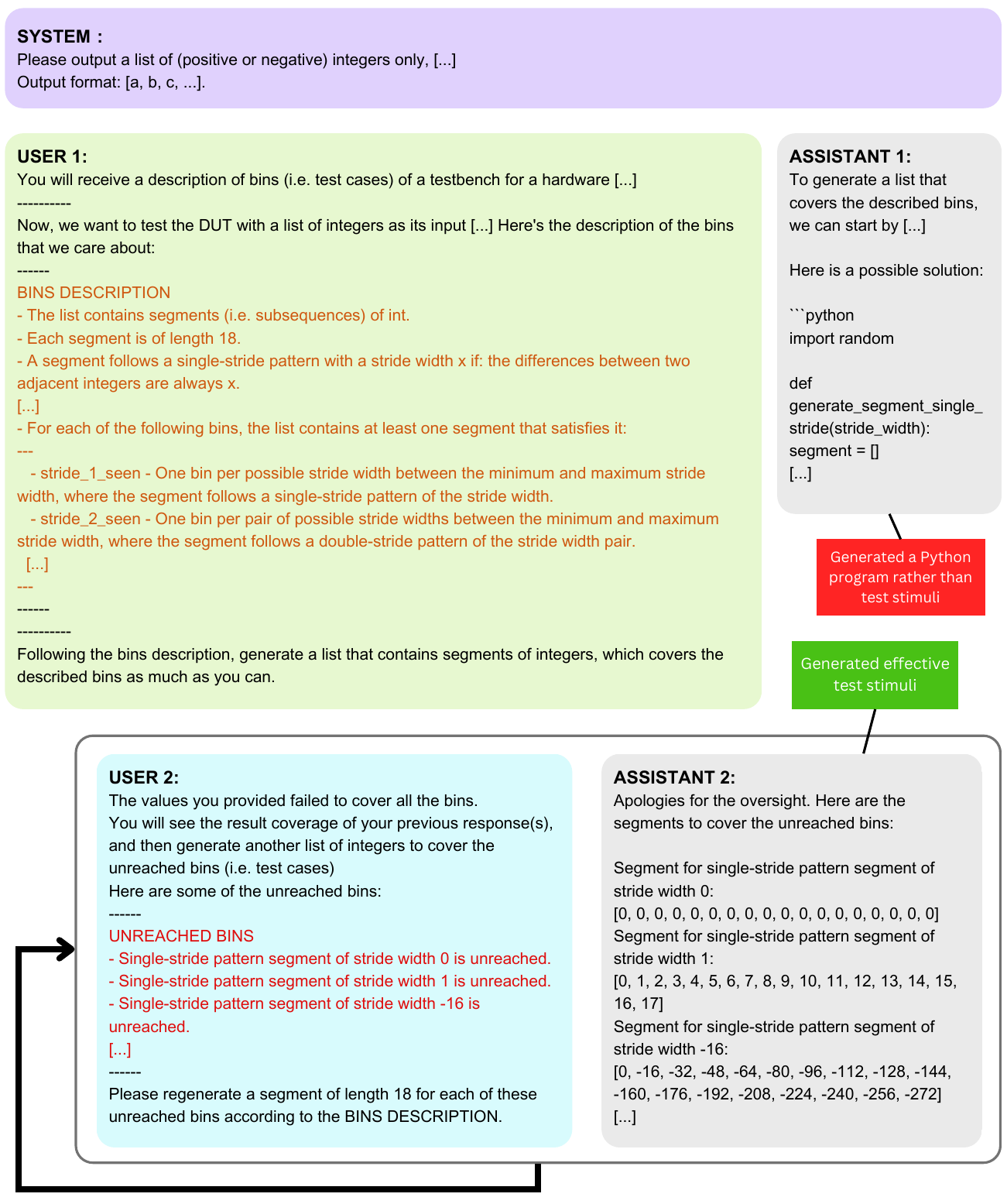}
  \caption{Example prompts and responses on the Primitive Data Prefetcher Core module. The purple box is the system message, containing a general format instruction. The green box is an initial query, containing a coverage plan summary (orange text). The blue box is an interactive query, containing differences i.e.\ coverage feedback (red text).
  }
  \label{example_dialogue}
\end{figure}

\textbf{Providing the DUT code} We include the DUT source code in the initial message to allow the LLM to leverage the HDL code to directly correlate specific features and functions with the corresponding coverage bins. Due to limited context length, this technique is applicable only when the device's code fits within the LLM's context window.


\textbf{Few-shot prompting} As task-specific fine-tuning is outside the scope of this study, we employ few-shot prompting to improve coverage metrics. Examples of stimuli with their corresponding bin hits are included in the initial prompt 
to facilitate the LLMs 
to assimilate the hardware verification context and the DUT information. To avoid skewing the experimental results, this technique is applicable only when the coverage plan includes more than 30 coverage bins.

\textbf{Example} Figure \ref{example_dialogue} demonstrates the LLM4DV flow works with several prompts and responses. The agent (USER) introduces the task and coverage plan in the initial message, and then provides coverage feedback in iterative messages. The LLM (ASSISTANT) generates textual responses according to the description and feedback.




\begin{table}

\centering
  \caption{Best results achieved for each LLM-DUT pair. Each reported experiment employs the best applicable configuration of the prompting techniques described in Section \ref{sec:experiments}. 
  Experiments marked with $^*$ have used few-shot prompting, and experiments marked with $^\dagger$ have included the DUT source code in the initial prompt. We highlight the \best{best} results for each DUT. Note that trials were limited to 700 messages.}
  \label{final_results}
  \resizebox{\linewidth}{!}{%
  \begin{tabular}{llrrrr}
    \toprule
    & & \makecell[c]{Primitive Data \\ Prefetcher Core}  & \makecell[c]{Asynchronous \\ FIFO}  & \makecell[c]{AMPLE Prefetcher \\ Weight Bank}  & \makecell[c]{AMPLE Prefetcher \\ Fetch Tag}  \\
    \midrule
    \multirow{3}{*}{gpt-3-turbo} & Max coverage & 1016 (98.26\%)$^*$  & 10 (100\%)  & 324 (100\%)$^\dagger$  & \best{10 (100\%)}  \\
     & Eff. msg. count & 350  & 16 & 36 & \best{2} \\
     & Avg. msg. count & 509.0 $\pm$ 129.4  &  19.7$\pm$3.9 &  37.7$\pm$1.2 & \best{22.0$\pm$14.1}  \\
     \midrule
     \multirow{3}{*}{llama-2-70b-chat} & Max coverage & 431 (41.68\%)$^*$  & 10 (100\%)$^\dagger$ & 324 (100\%) & 10 (100\%) \\
     & Eff. msg. count & 700  & 1 & 36 & 22 \\
     & Avg. msg. count &  470.7$\pm$189.9 & 10.5$\pm$7.9 & 41.3$\pm$7.5 & 27.7$\pm$6.0 \\
     \midrule
     \multirow{3}{*}{claude-3-sonnet} & Max coverage & 801 (77.47\%)$^*$ & \best{10 (100\%)} & 324 (100\%) & 10 (100\%) \\
     & Eff. msg. count & 700  & \best{1} & 36 & 8 \\
     & Avg. msg. count & 676.3$\pm$33.5 & \best{1.0} & 36.0 & 19.3$\pm$8.0 \\
     \midrule
     \multirow{3}{*}{codellama-70b-instruct} & Max coverage & 82 (7.93\%)$^*$ & 10 (100\%) & 324 (100\%) & 6 (60.00\%)  \\
     & Eff. msg. count & 154 & 1 & 44 & 34 \\
     & Avg. msg. count & 102.0$\pm$50.3 & 3.7$\pm$3.1 & 52.3$\pm$8.5 & 28.3$\pm$4.0  \\
     \midrule
     \multirow{3}{*}{llama-3-70b-instruct} & Max coverage & 710 (68.67\%)$^*$ & 10 (100\%)$^\dagger$ & \best{324 (100\%)} & 10 (100\%) \\
     & Eff. msg. count & 700 & 1 & \best{26} & 15 \\
     & Avg. msg. count & 700.0 & 1.3$\pm$0.5 & \best{32.7$\pm$4.7} & 20.0$\pm$3.6 \\
     \midrule
     \multirow{3}{*}{claude-3.5-sonnet} & Max coverage & \best{1022 (98.84\%)$^*$} & \best{10 (100\%)} & 324 (100\%) & 9 (90\%) \\
     & Eff. msg. count & \best{321} & \best{1} & 36 & 25 \\
     & Avg. msg. count & \best{329.3$\pm$32.3} & \best{1.0} & 36.7$\pm$0.6 & 25.0 \\
     \midrule
     Formal verification & Max coverage & 1030 (99.61\%) & 10 (100\%)  & 3 (0.93\%) & 10 (100\%) \\
     \midrule
     CRT & Max coverage & 0 (0\%) & 10 (100\%) & 324 (100\%) & 10 (100\%) \\
     \bottomrule
\end{tabular}
}
\resizebox{\linewidth}{!}{%
\begin{tabular}{llrrrr}
\toprule
    & & \makecell[c]{SDRAM \\ Controller} & \makecell[c]{Ibex CPU \\ Instruction Decoder} & \makecell[c]{Ibex CPU} & \makecell[c]{MIPS CPU} \\
    \midrule
     \multirow{3}{*}{gpt-3-turbo} & Max coverage & 7 (100\%) & 1466 (69.58\%)$^*$  & 39 (19.90\%)$^*$  & 84 (43.08\%)$^*$  \\
     & Eff. msg. count & 7  & 700 & 102 & 211 \\
     & Avg. msg. count & 22.3$\pm$11.0  & 432.0$\pm$228.3  & 88.0$\pm$21.2 & 111.0$\pm$72.8 \\
     \midrule
     \multirow{3}{*}{llama-2-70b-chat} & Max coverage & 6 (85.71\%)$^\dagger$ & 402 (19.08\%)$^*$  & 22 (11.22\%)$^*$ & 68 (34.87\%)$^*$  \\
     & Eff. msg. count & 32 & 186 & 26 & 55 \\
     & Avg. msg. count & 28.3$\pm$2.6  & 125.7$\pm$61.1 & 33.3$\pm$10.4 & 45.7$\pm$13.2 \\
     \midrule
     \multirow{3}{*}{claude-3-sonnet} & Max coverage & 7 (100\%)$^\dagger$  & 1512 (71.76\%)$^*$ & 141 (71.94\%)$^*$ & 159 (81.54\%)$^*$ \\
     & Eff. msg. count & 2  & 700 & 315 & 299 \\
     & Avg. msg. count & 2.3$\pm$0.5 & 700.0 & 287$\pm$19.9 & 277.7$\pm$35.2 \\
     \midrule
     \multirow{3}{*}{codellama-70b-instruct} & Max coverage & 7 (100\%)$^\dagger$ & 417 (19.79\%)$^*$ & 25 (12.76\%)$^*$ & 91 (46.67\%)$^*$ \\
     & Eff. msg. count & 8 & 182 & 31 & 142 \\
     & Avg. msg. count & 29.3$\pm$15.1 & 126.3$\pm$57.6 & 34.3$\pm$6.9 & 113.7$\pm$20.4 \\
     \midrule
     \multirow{3}{*}{llama-3-70b-instruct} & Max coverage & \best{7 (100\%)} & 1135 (53.89\%)$^*$ & 94 (47.96\%)$^*$ & 98 (50.26\%)$^*$ \\
     & Eff. msg. count & \best{1} & 700 & 172 & 175 \\
     & Avg. msg. count & \best{2.3$\pm$1.2} & 700 & 180.3$\pm$20.9 & 141$\pm$24.1 \\
     \midrule
     \multirow{3}{*}{claude-3.5-sonnet} & Max coverage & 7 (100\%)$^\dagger$ & \best{2006 (95.21\%)$^*$} & \best{196 (100\%)$^*$} & \best{175 (89.74\%)$^*$} \\
     & Eff. msg. count & 2 & \best{651} & \best{31} & \best{176} \\
     & Avg. msg. count & 2.0 & \best{683.7$\pm$28.3} & \best{37.0$\pm$5.29} & \best{174.7$\pm$41.0} \\
     \midrule
     Formal verification & Max coverage &  7 (100\%) & 2106 (99.95\%) & 100\% & 100\% \\
     \midrule
     CRT & Max coverage &  7 (100\%) & 1154 (54.77\%) & 30 (15.31\%) & 28 (14.36\%) \\
    \bottomrule
  \end{tabular}
}  
\end{table}

\section{Results and Analysis}
\label{sec:experiments}
Table \ref{final_results} presents the best results achieved for each LLM-DUT pair, compared with naive CRT and formal methods serving. In the CRT methodology, we generate 100,000 combinations within the valid input range without additional constraints.
The formal baseline utilizes the cover mode of the SymbiYosys tool \cite{yosys}, where all bins of the coverage plans correspond to specific SystemVerilog cover statements, and each formal verification run is limited to a 48-hour timeout.

Across all DUTs, each configuration demonstrates that LLM4DV can either match or exceed the coverage rates achieved via naive CRT. This signifies not only the adaptability of LLMs to varied hardware testing contexts but also their potential to streamline certain aspects of verification by reducing reliance on extensive random input generation.

\section{Conclusion: gimmick or trend?}
The hardware design community is now starting to see debates regarding the effectiveness of LLMs for automated chip design, questioning whether their use is merely a gimmick or represents a future trend. 
Our particular take on this problem is that there is a need to set up open datasets and benchmarks for problems in chip design, so that the effectiveness and potential use of LLMs can be fully understood and quantified. 

LLM4DV fits exactly in this category, and we target, in our opinion, the most human labor-intensive part (in terms of engineering) of the chip design process.
Our baseline results have demonstrated that LLMs can achieve satisfactory coverage rates on straightforward designs, but they struggle with more complex ones, suggesting that LLMs do hold promise within the specific context of automated hardware DV.

\balance
\bibliographystyle{IEEEtran}
\bibliography{references}


\end{document}